\documentclass[10pt,twocolumn,letterpaper]{article}

\usepackage{cvpr}              

\usepackage{graphicx}
\usepackage{amsmath}
\usepackage{amssymb}
\usepackage{booktabs}
\usepackage{multirow}
\usepackage[pagebackref,breaklinks,colorlinks]{hyperref}

\usepackage[capitalize]{cleveref}
\crefname{section}{Sec.}{Secs.}
\Crefname{section}{Section}{Sections}
\Crefname{table}{Table}{Tables}
\crefname{table}{Tab.}{Tabs.}

\def\cvprPaperID{11806}
\def\confName{ICCV}
\def\confYear{2023}

\begin{document}

\title{FalconNet: Factorization for the Light-weight ConvNets}

\author{Zhicheng Cai\\
Nanjing University\\
{\tt\small caizc@smail.nju.edu.cn}
\and
Qiu Shen\\
Nanjing University\\
{\tt\small shenqiu@.nju.edu.cn}
}
\maketitle

\begin{abstract}
Designing light-weight CNN models with little parameters and Flops is a prominent research concern. However, three significant issues persist in the current light-weight CNNs: i) the lack of architectural consistency leads to redundancy and hindered capacity comparison, as well as the ambiguity in causation between architectural choices and performance enhancement; ii) the utilization of a single-branch depth-wise convolution compromises the model representational capacity; iii) the depth-wise convolutions account for large proportions of parameters and Flops, while lacking efficient method to make them light-weight. To address these issues, we factorize the four vital components of light-weight CNNs from coarse to fine and redesign them: i) we design a light-weight overall architecture termed LightNet, which obtains better performance by simply implementing the basic blocks of other light-weight CNNs; ii) we abstract a Meta Light Block, which consists of spatial operator and channel operator and uniformly describes current basic blocks; iii) we raise RepSO which constructs multiple spatial operator branches to enhance the representational ability; iv) we raise the concept of receptive range, guided by which we raise RefCO to sparsely factorize the channel operator. Based on above four vital components, we raise a novel light-weight CNN model termed as FalconNet. Experimental results validate that FalconNet can achieve higher accuracy with lower number of parameters and Flops compared to existing light-weight CNNs.
\end{abstract}

\section{Introduction}
Convolutional Neural Networks (CNNs) possess the capability of representing high-dimensional complex functions and have been successfully applied to various real visual scenarios~\cite{krizhevsky2012imagenet,simonyan2014very, he2016deep,huang2017densely}. With lower computational complexity and higher efficiency than ViTs, CNNs remain the  dominance in computer vision applications~\cite{liu2022convnet,yu2023inceptionnext,ding2022scaling}.
For the implementation on mobile devices for real-world applications, the computational and storage resources are always limited, requiring light-weight CNN models with reduced parameters and low Flops while maintaining competitive performance.
Depth-wise separable convolution (DS-Conv), proposed in MobileNetV1~\cite{howard2017mobilenets}, factorizes the regular convolution into depth-wise convolution (DW-Conv) and point-wise convolution (PW-Conv), which extracts the spatial and channel features individually. Consequently, DS-Conv decreases a large amount of computation and parameters and has been a fundamental design component for subsequent light-weight CNNs~\cite{sandler2018mobilenetv2,howard2019searching,zhou2020rethinking,li2019hbonet,tan2019mnasnet,tan2019efficientnet,tan2021efficientnetv2,zhang2023rethinking} and modern large CNNs~\cite{liu2022convnet,yu2023inceptionnext,ding2022scaling}. 
MobileNetv2~\cite{sandler2018mobilenetv2} proposes the inverted residual block (IRB) to alleviate the destruction to the features and enhance representational capacity. Utilizing the paradigm of IRB as the basic block, many light-weight models with different overall architectures (stages, width, depth)~\cite{howard2019searching,tan2019mnasnet,tan2019efficientnet,tan2021efficientnetv2,han2020model,wu2019fbnet} are raised. In addition to IRB, many efficient basic blocks~\cite{tan2019mixconv,zhou2020rethinking,li2019hbonet,zhang2023rethinking,chen2023run} with different structures are designed to improve the representational ability of light-weight CNNs. However, the architectural inconsistencies cause redundant structures that could be consolidated through unification, and the use of varying basic blocks along with varying architectures leading to unfair capacity comparisons and obscuring the causal relationship between architectural choices and performance enhancements. 

\begin{figure}[t]
  \centering
  \includegraphics[width=.46\textwidth]{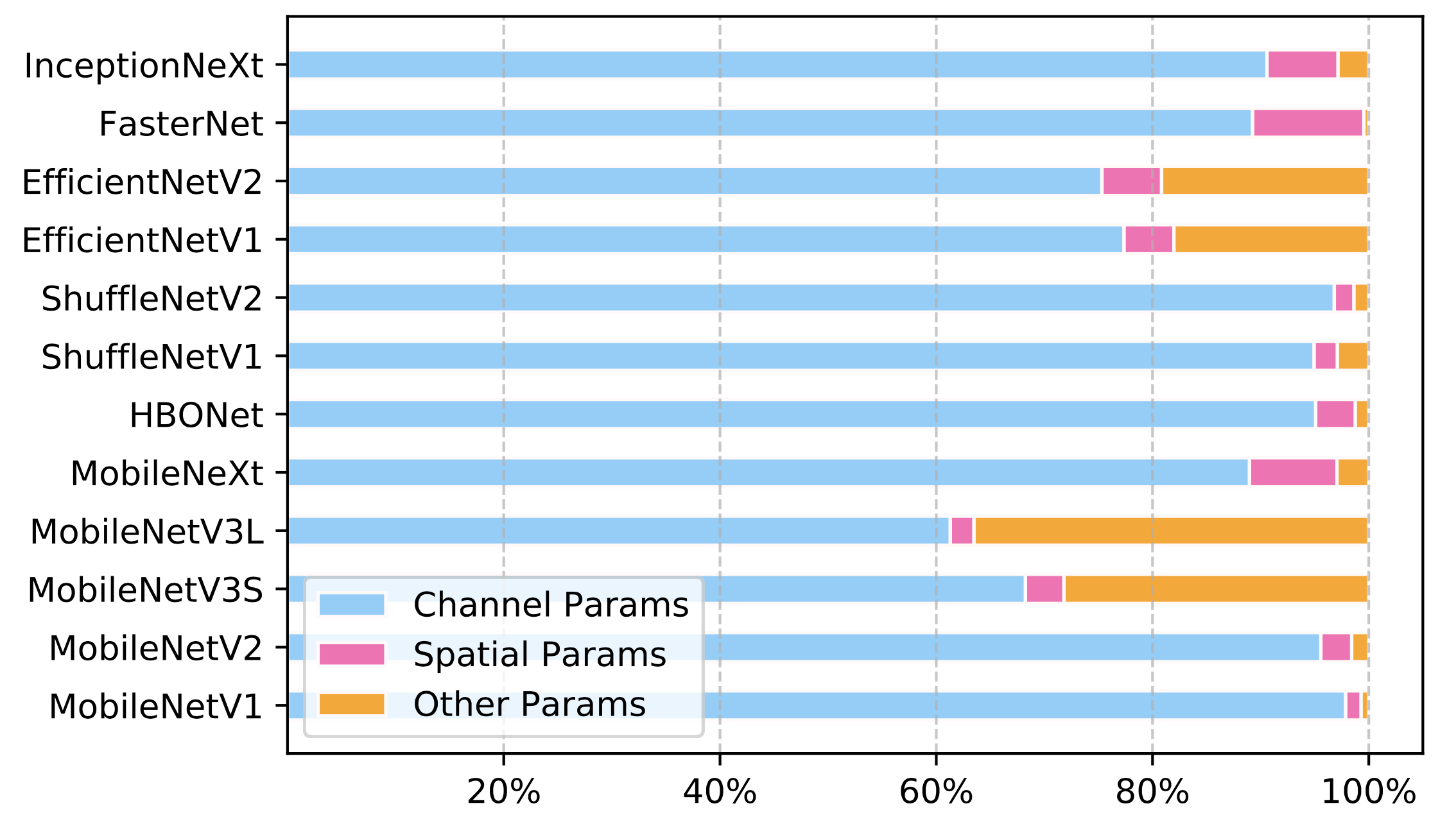}
  \caption{The parameter proportions of channel parameters, spatial parameters and other parameters (ignore the classifier head). }
  \label{p1}
  \vspace{-1em} 
\end{figure}

Moreover, while some works~\cite{szegedy2016rethinking,li2021micronet,yu2023inceptionnext,liu2022more,tang2022ghostnetv2} factorize the DW Conv into parallel low-rank branches to save computational cost, the main stem for light-weight models chasing for even lower Flops and parameters lies in the PW Conv.
As Fig.~\ref{p1} exhibited, the PW Conv (broadly, the densely-connected linear layers processing the channel information) accounts for the majority of the light-weight model parameters, while the DW Conv (broadly, conv layers processing the spatial information) only makes up a small proportion. 
ShuffleNet splits channels into groups and shuffles channels~\cite{zhang2018shufflenet,ma2018shufflenet,li2021micronet}, however, the information communication between channels in different groups is insufficient.   
IGC~\cite{zhang2017interleaved,xie2018interleaved,sun2018igcv3} raises interleaved group convolutions to replace the regular convolution with multiple permuted group convolution layers, while the structure becomes complicated.
ChannelNet~\cite{convolutions2021channelnets} introduces sparse connections to PW-Conv, while one output channel can only attends to a small fraction of the input channels, thus is only applied to the last PW-Conv layer due to the inadequate information communication. 

This paper want to raise a novel light-weight CNN model with small amount of parameters while maintaining competitive performance. In light of the aforementioned issues, we factorize the four vital components of constructing light-weight CNNs, namely, overall architecture, meta basic block, spatial operator and channel operator. 
To be specific:  
\begin{itemize}
\item We first design a light-weight overall architecture termed as \textbf{\emph{LightNet}}, which refers to the structural designs of modern CNNs and has four stages, each stage is stacked with basic blocks. Better performance can be obtained by simply implementing the basic blocks of other light-weight CNNs on LightNet.
\item We abstract and analyze a \textbf{\emph{Meta Basic Block}}, consisting of spatial operator and channel operator (specifically, PW-Conv), for light-weight CNN model design. The paradigm of meta basic block uniformly describes the current basic blocks, e.g., IRB in MobileNets~\cite{sandler2018mobilenetv2,howard2019searching} and EfficientNets~\cite{tan2019mnasnet,tan2019efficientnet,tan2021efficientnetv2,han2020model,wu2019fbnet}, sandglass block~\cite{zhou2020rethinking} and FasterNet block~\cite{chen2023run}, inferring that the framework of the meta block provides the basic ability to the model, while the differences of model performances essentially come from different structural instantiations~\cite{zhang2023rethinking}. Through extensive experiments we further simplify meta basic block into \textbf{\emph{Meta Light Block}}, which obtains better performance. 
\item We introduce \textbf{\emph{RepSO}} as the spatial operator for the Meta Light block. According to the guidance of weight magnitude, RepSO constructs multiple extra branches to compensate for the reduce of learnable parameters and enhance the model representational capacity. RepSO further utilizes the structural reparameterization methodology to equivalently convert these diverse branches into a single branch in inference.  
\item We introduce the concept of \textbf{\emph{Receptive Range}} for channel dimension correspondence to the concept of receptive field for spatial dimension.   
Receptive range elaborates the way of connection between the output and input neurons in the PW-Conv, which claims that one output neuron should attend to all of the input neurons directly or indirectly (first attend to a set of hidden neurons which attend to all the input neurons) to obtain the full receptive range. 
Based on the concept of receptive range we further raise \textbf{\emph{RefCO}} which factorizes the PW-Conv in the Meta Light block through introducing sparsity to the dense channel connection correspondence to the spatial convolution. Moreover, RefCO utilizes structural reparameterization while construct multiple \textbf{\emph{sparsely factorized PW-Convs}} to compensate for the reduction of channel connections.
\item Finally, we raise a novel light-weight CNN model termed as \textbf{\emph{FalconNet}} based on above four vital components. Experimental results show that FalconNet can achieve higher accuracy with less parameters and Flops compared to existing light-weight CNNs.
\end{itemize}
\begin{figure*}[t]
  \centering
  \includegraphics[width=.96\textwidth]{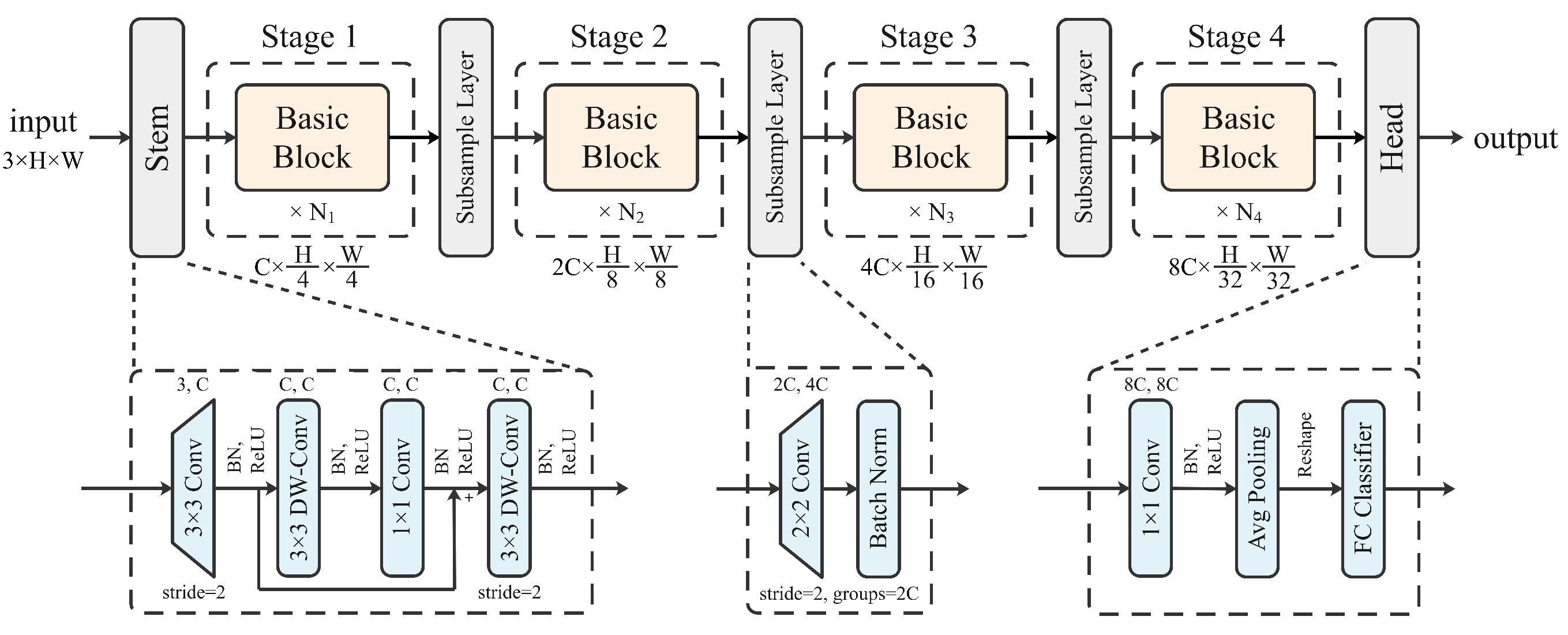}
  \caption{LightNet overall architecture. }
  \label{p2}
  \vspace{-1em} 
\end{figure*}

\section{Related Works}
\subsection{Light-weight CNN Models}
In order to deploy on mobile devices for real-world applications, many light-weight CNN models with reduced parameter amounts and limited computational burdens are proposed.
SqueezeNet~\cite{iandola2016squeezenet} raises fire module which partially replaces the $3\times3$ convolution kernels with $1\times1$ kernels. 
InceptionV3~\cite{szegedy2016rethinking} factorizes the standard convolution into two asymmetric convolutions. 
IGC~\cite{zhang2017interleaved,xie2018interleaved,sun2018igcv3} raises interleaved group convolutions to decompose the regular convolution into multi-layer group convolutions. 
ShuffleNetV1~\cite{zhang2018shufflenet} and MicroNet~\cite{li2021micronet} utilizes $1\times1$ group convolutions and then shuffles the grouped channels, while ShuffleNetV2~\cite{ma2018shufflenet} firstly splits the channels then shuffles the channels. 
ChannelNet~\cite{convolutions2021channelnets} introduces sparse connections to the dense layers. 
MobileNetV1~\cite{howard2017mobilenets} and Xception~\cite{chollet2017xception} propose the depth-wise separable convolution to decouple the regular convolution into depth-wise convolution and point-wise convolution, which alleviates a large amount of computation and parameters and has been a widely-adopted design element for modern efficient CNN models~\cite{liu2022convnet,yu2023inceptionnext,ding2022scaling}. 
MobileNetV2~\cite{sandler2018mobilenetv2} introduces the inverted residual block. 
MobileNetV3~\cite{howard2019searching} enhances MobileNetV2 with squeeze-and-excitation module~\cite{hu2018squeeze} and neural architecture search~\cite{zoph2016neural,liu2018darts,zoph2018learning}. 
MobileNeXt~\cite{zhou2020rethinking} introduces sandglass block to alleviate information loss by flipping the structure of inverted residual block. 
HBONet~\cite{li2019hbonet} raises harmonious bottleneck on two orthogonal dimensions to improve representation.
EfficientNet~\cite{tan2019mnasnet,tan2019efficientnet,tan2021efficientnetv2,wu2019fbnet} proposes a compound scaling method to scale depth, width and resolution uniformly. 
EMO~\cite{zhang2023rethinking} additionally introduces the self-attention to the inverted residual block. 
FasterNet~\cite{chen2023run} raises partial convolution to conduct regular convolution on part of the channels to reduce Flops while increasing FLOPS.

\subsection{Structural Reparameterization}
Structural Reparameterization~\cite{ding2019acnet,ding2021resrep,ding2021diverse,ding2021repvgg,ding2022repmlpnet,ding2022scaling} is a representative reparameterization methodology to parameterize a structure with the parameters transformed from another structure. 
Typically, it adds extra branches to the model in training to enhance the representational capacity and improve the performance, then equivalently simplifies the training structure into the same as the original model for inference, without any extra computational or memory cost.  
ACNet~\cite{ding2019acnet} asymmetrically constructs two extra vertical and horizontal convolution branches in training and converts them into the original branch in inference. 
RepVGG~\cite{ding2021repvgg} constructs identity shortcuts parallel to the $3\times3$ convolution during training and converts the shortcuts into the $3\times 3$ branches. 
DBB~\cite{ding2021diverse} constructs inception-like diverse branches of different scales and complexities to enrich the feature space.
RepLKNet~\cite{ding2022scaling} adds a relatively small kernel into the large kernel to capture small-scale patterns.
MobileOne~\cite{vasu2022improved} also multiplies the convolution branch in training and reparameterizes them in inference for performance improvement.

\section{Method}
\subsection{Design the Overall Architecture}
Firstly, although certain light-weight CNNs utilize similar basic building blocks, their overall architectures differ, resulting in unnecessary redundancy that could be consolidated through unification.
Secondly, certain light-weight CNNs incorporate distinct basic building blocks, yet they still exhibit varying overall architectures. This leads to an unfair comparison of capacity between the basic blocks and obscures the causal relationship between architectural choices and performance improvements.
Hence we first intend to construct an overall architecture especially for light-weight CNN models. Referring to the modern architectures of powerful CNN~\cite{chen2023run,ding2022scaling,liu2022convnet,yu2023inceptionnext} and ViT~\cite{liu2021swin,liu2022swin,dong2022cswin} models, we raise \textbf{\emph{LightNet}} of which architecture is sketched in Fig.~\ref{p2}.

\textbf{Stem.} Stem refers to the beginning part of the model. Instead of utilizing single conv layer with relatively large stride, we desire to capture more details by several conv layers at the beginning. After the first $3\times3$ regular conv layer with a stride of 2 , we employ a $3\times3$ DW-Conv layer to capture low-level patterns, followed by a $1\times1$ PW-Conv layer and another $3\times3$ DW-Conv layer for subsampling~\cite{ding2022scaling}. There also exists a shortcut as shown in Fig.~\ref{p2}.

\textbf{Stages.} Stages 1-4 are composed of several repeated \emph{basic blocks}, such as IRB. According to the stage compute ratio of 1:1:3:1~\cite{liu2022convnet,liu2021swin}, the number of blocks in each stage is set as [3, 3, 9, 3]. Following the pyramid principle~\cite{he2016deep} and considering reducing the parameters, the channel dimension in each stage is set as [32, 64, 128, 256]. 

\textbf{Subsampling Layers.} We add separate subsampling layers between stages. We use $2\times2$ conv layer with groups of input channel dimension and a stride of 2 for halving the spatial resolution. The $2\times2$ conv layer also doubles the channel dimension. A batch normalization layer is arranged subsequently to stablize training.

\textbf{Head.} Head, following the 4-th stage, is the last part of the model. We first use a $1\times1$ conv layer to further mix the information, then utilize the global average pooling to obtain the feature vectors, which is subsequently input into the last fully-connected classifier and obtain the final output. 

Experimental results show that simply implementing the basic blocks of other light-weight CNNs on LightNet achieves better performance than the original models.

\begin{figure}[tbp]
  \centering 
  \begin{subfigure}{0.24\linewidth}
    \includegraphics[width=0.99\linewidth]{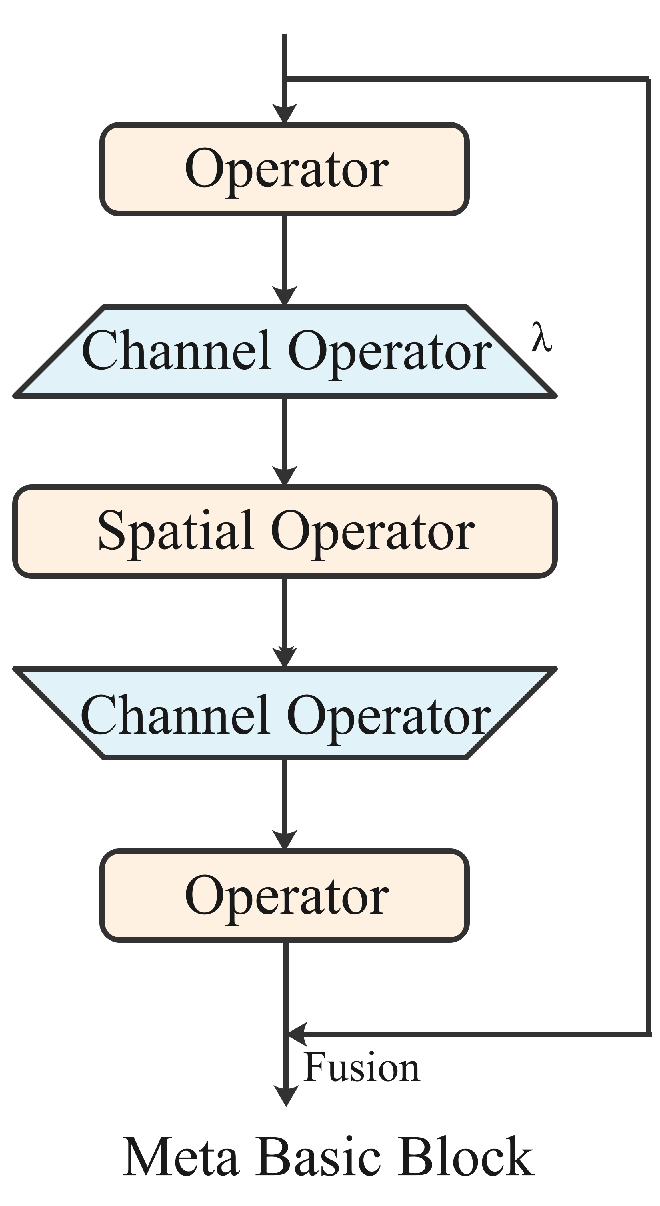}
    \caption{ }
    \label{p3-1-1}
  \end{subfigure}
  \begin{subfigure}{0.24\linewidth}
    \includegraphics[width=0.99\linewidth]{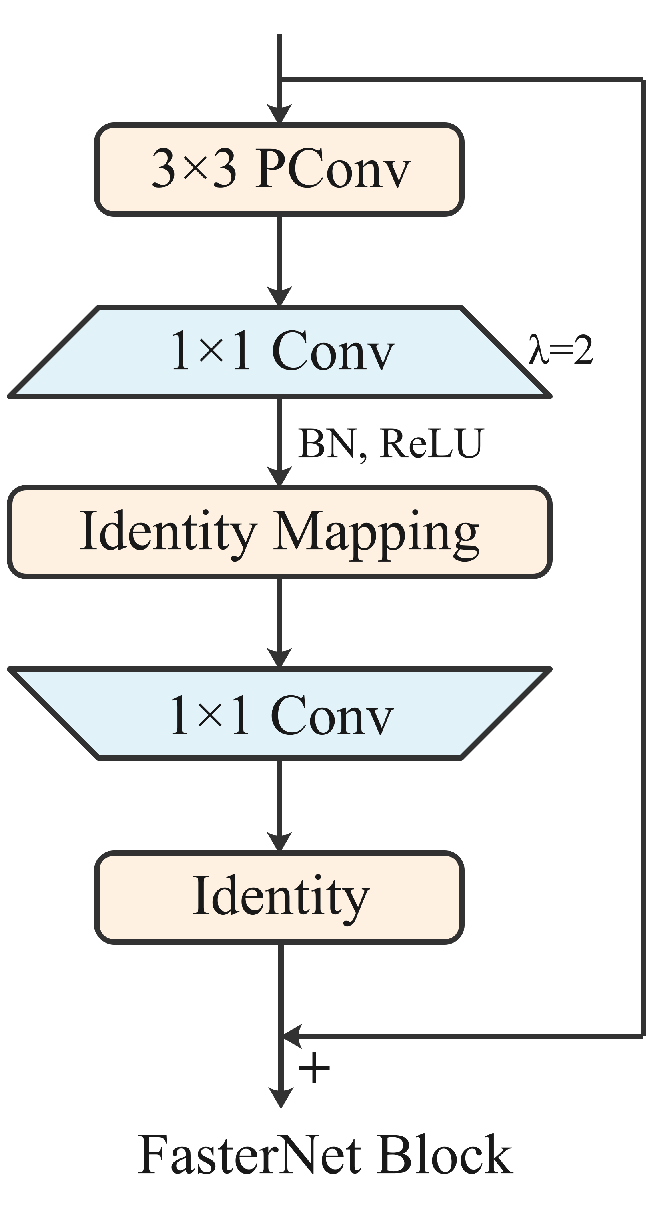}
    \caption{ }
    \label{p3-1-2}
  \end{subfigure}
  \begin{subfigure}{0.24\linewidth}
    \includegraphics[width=0.99\linewidth]{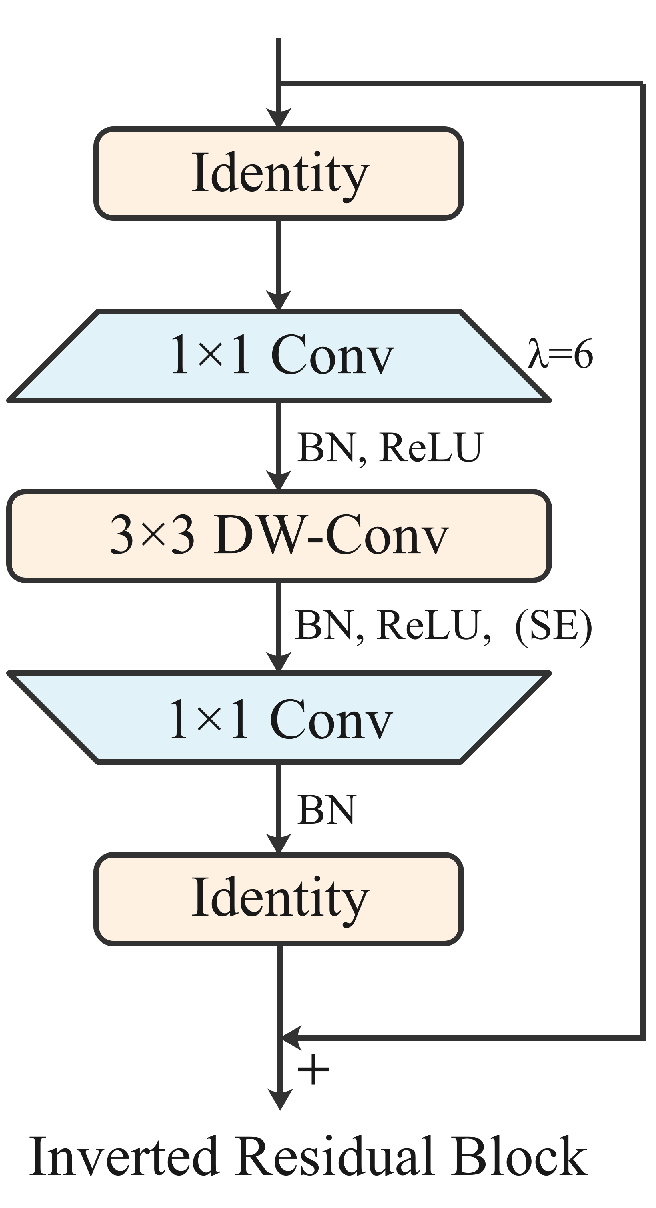}
    \caption{ }
    \label{p3-1-3}
  \end{subfigure}
  \begin{subfigure}{0.24\linewidth}
    \includegraphics[width=0.99\linewidth]{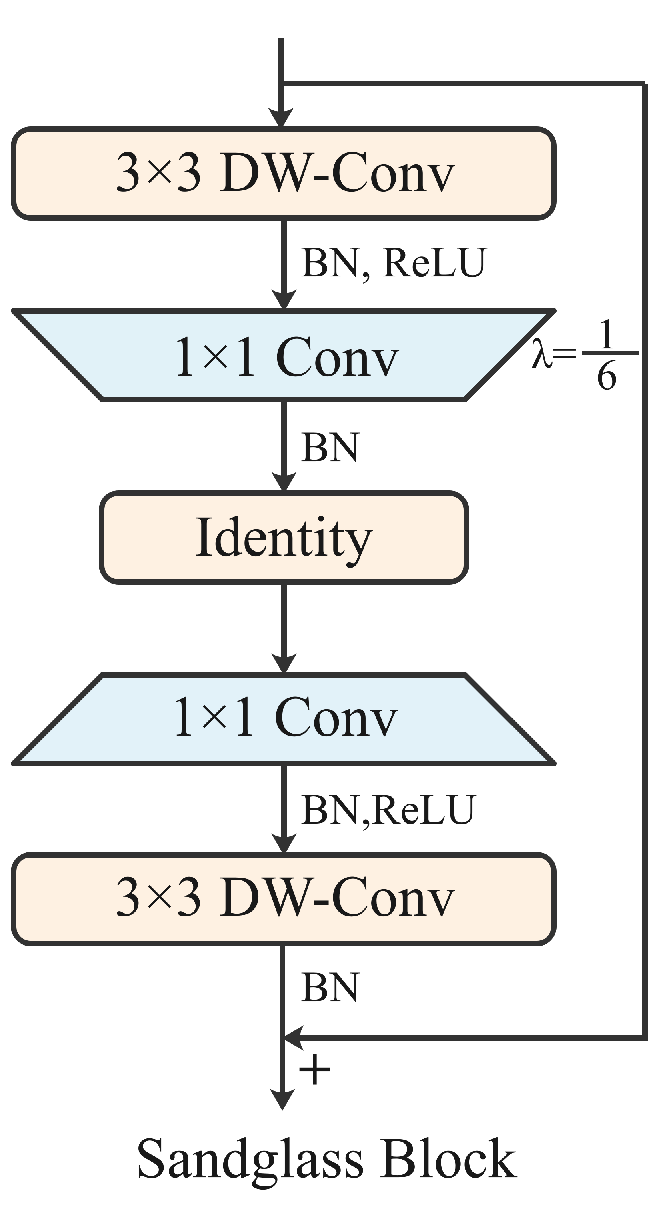}
    \caption{ }
    \label{p3-1-4}
  \end{subfigure}
  \caption{Abstracted unified Meta Basic Block for light-weight models. Some correspondingly instantiated basic blocks (e.g., IRB and sandglass block) are also selected to exhibit.}
  \label{p3-1}
  \vspace{-1em} 
\end{figure}

\begin{figure}[tbp]
  \centering 
  \begin{subfigure}{0.24\linewidth}
    \includegraphics[width=0.99\linewidth]{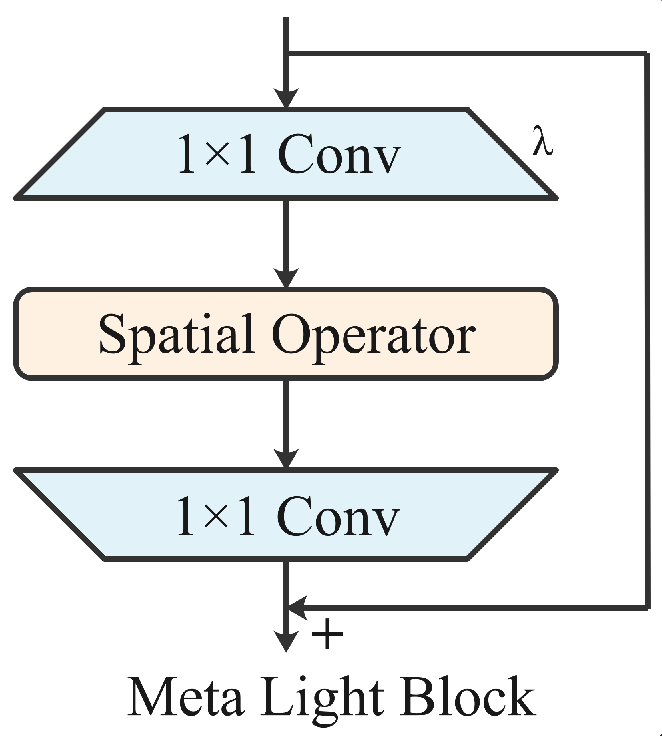}
    \caption{ }
    \label{p3-2-1}
  \end{subfigure}
  \begin{subfigure}{0.24\linewidth}
    \includegraphics[width=0.99\linewidth]{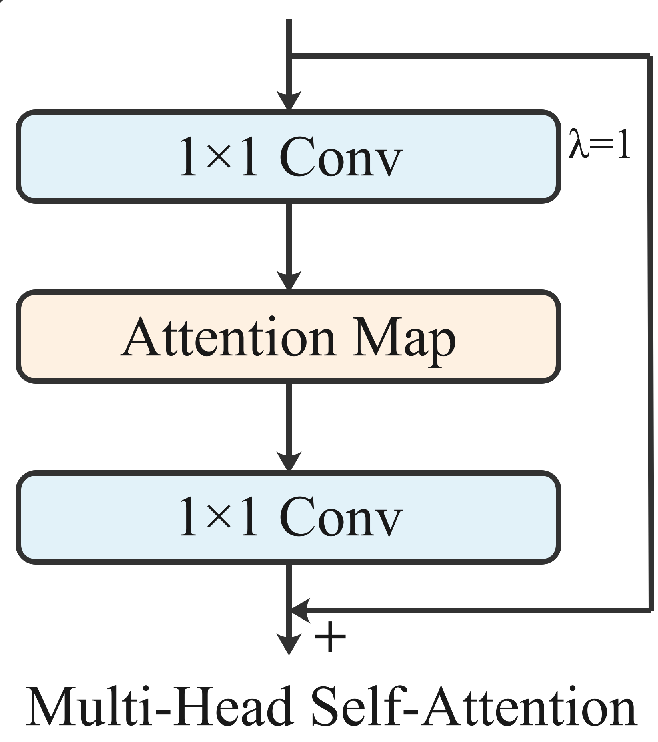}
    \caption{ }
    \label{p3-2-2}
  \end{subfigure}
  \begin{subfigure}{0.24\linewidth}
    \includegraphics[width=0.99\linewidth]{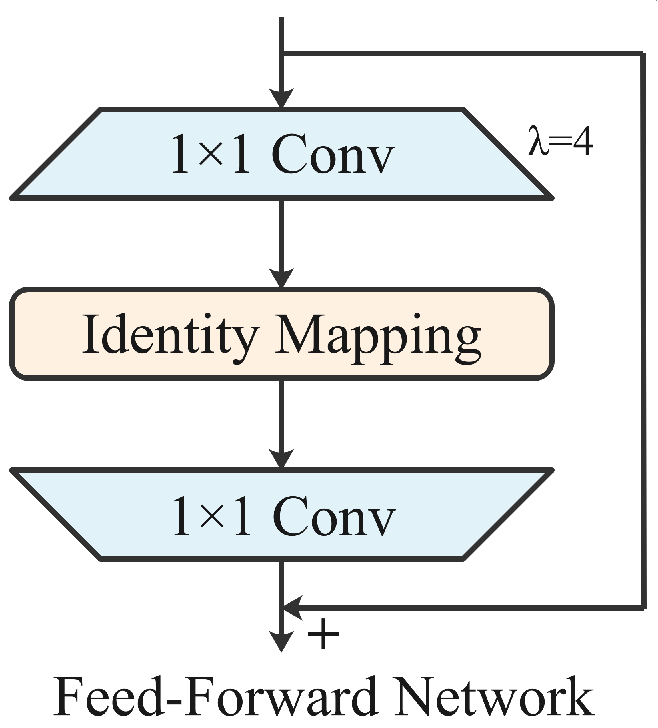}
    \caption{ }
    \label{p3-2-3}
  \end{subfigure}
  \begin{subfigure}{0.24\linewidth}
    \includegraphics[width=0.99\linewidth]{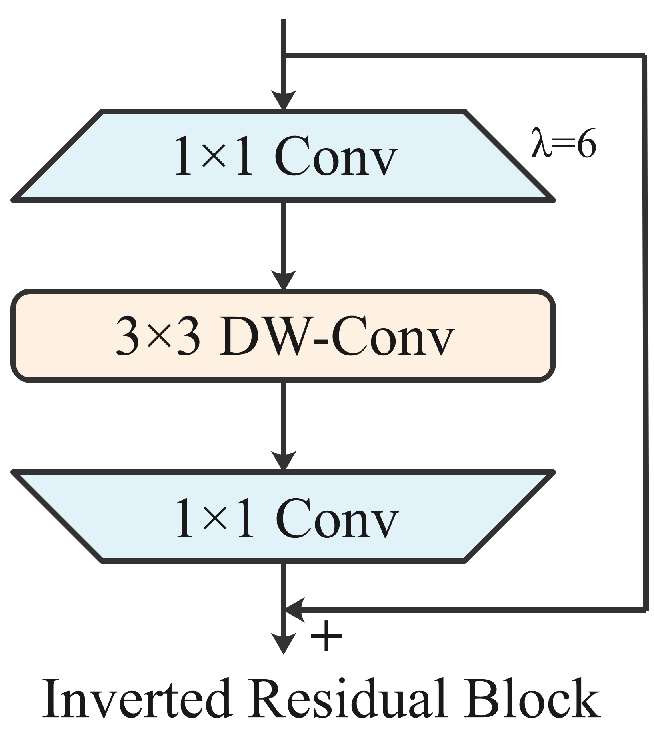}
    \caption{ }
    \label{p3-2-4}
  \end{subfigure}
  \caption{Simplified abstracted Meta Light Block for efficient light-weight models. Some correspondingly instantiated basic blocks (e.g., IRB and FFN) are also selected to exhibit.}
  \label{p3-2}
  \vspace{-1em} 
\end{figure}
\subsection{Explore the Meta Basic Block}\label{mb}
Basic blocks are the pivotal component for light-weight CNNs. As exhibited in Fig.~\ref{p3-1}, different basic blocks can be generally abstracted into the \textbf{\emph{Meta Basic Block}} (Fig.~\ref{p3-1-1}), which is alternately composed of spatial and channel operators (i.e., PW-Conv). The framework of the meta basic block provides the basic ability to the model, which means instantiating these spatial operators as non-learnable identity mappings can still achieve effective performance, while the differences of model performances essentially come from different structural instantiations of the meta basic block, e.g., FasterNet block instantiates the first spatial operator as PConv and instantiates the other two spatial operators as identity mapping, as shown in Fig.~\ref{p3-1-1}. 
Through conducting extensive experiments (Sec.~\ref{exp-metablock}), it is observed that the first and the last operator layers make no benefit to the enhancement of model performance, while the second spatial operator layer between two channel operator layers is significant.
Thus we further simplify meta basic block into \textbf{\emph{Meta Light Block}} (Fig.~\ref{p3-2-1}), which consists of two PW-Conv layers (with an expansion ratio $\lambda$) and a single spatial operator layer in between. 

\begin{figure}[t]
  \centering 
  \begin{subfigure}{0.24\linewidth}
    \includegraphics[width=0.99\linewidth]{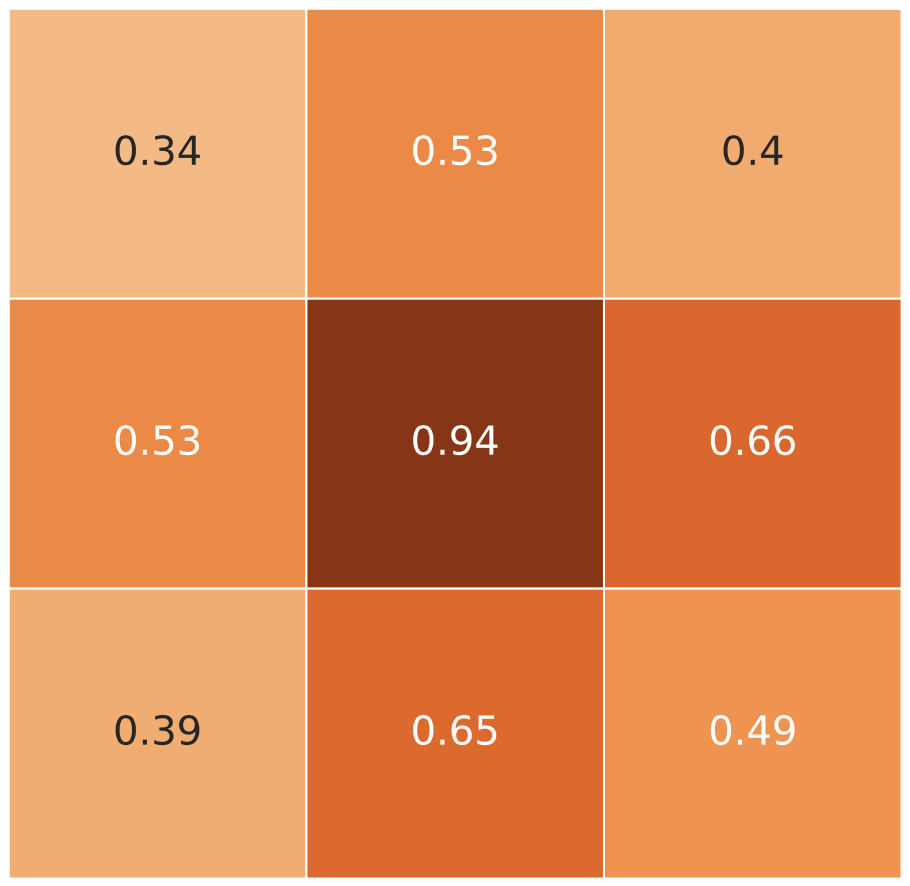}
    \caption{MobileNetV2}
    \label{p5-1}
  \end{subfigure}
  \begin{subfigure}{0.24\linewidth}
    \includegraphics[width=0.99\linewidth]{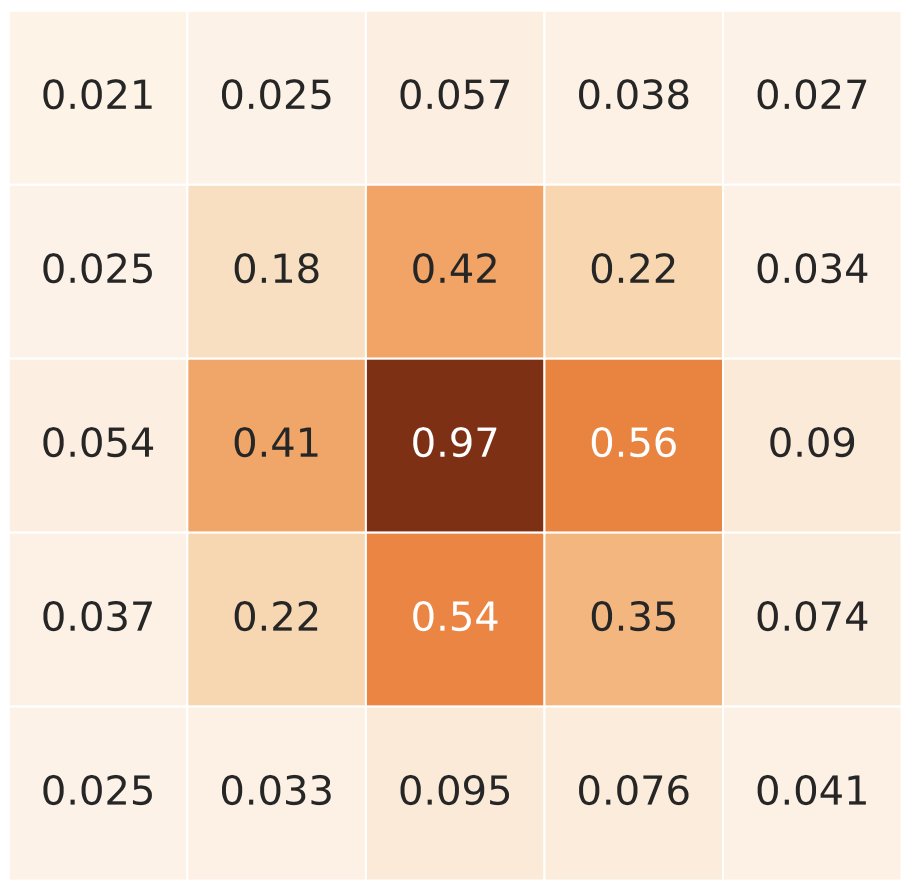}
    \caption{MobileNetV3}
    \label{p5-2}
  \end{subfigure}
  \begin{subfigure}{0.24\linewidth}
    \includegraphics[width=0.99\linewidth]{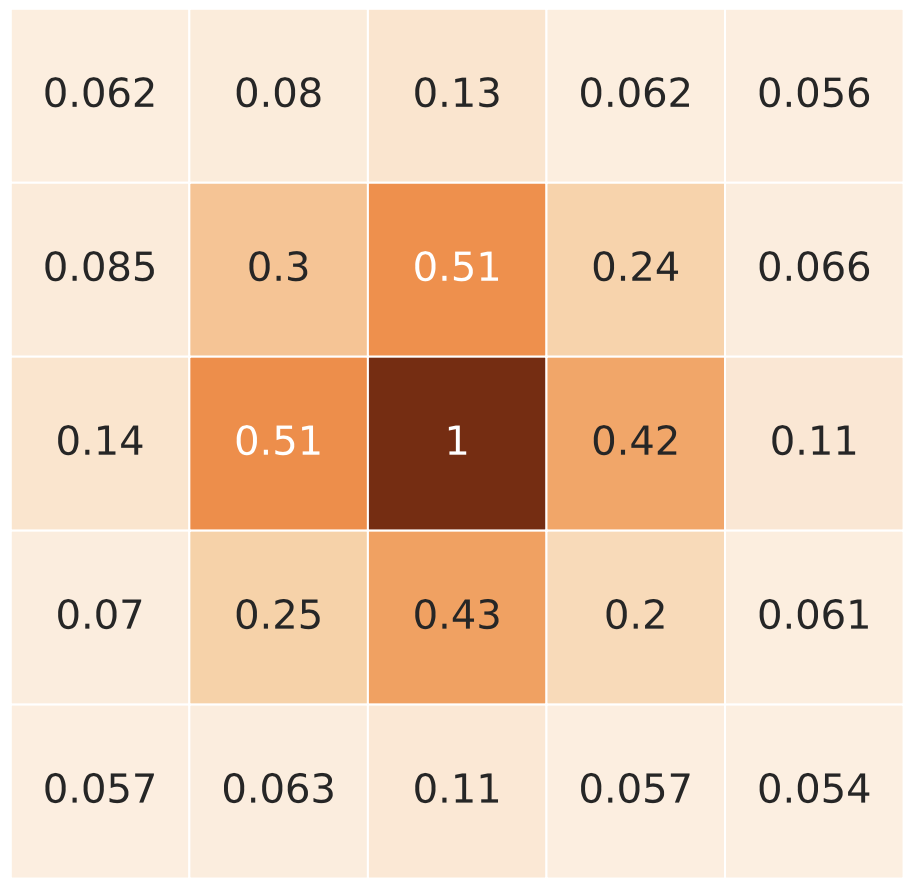}
    \caption{EfficientNet}
    \label{p5-3}
  \end{subfigure}
  \begin{subfigure}{0.24\linewidth}
    \includegraphics[width=0.99\linewidth]{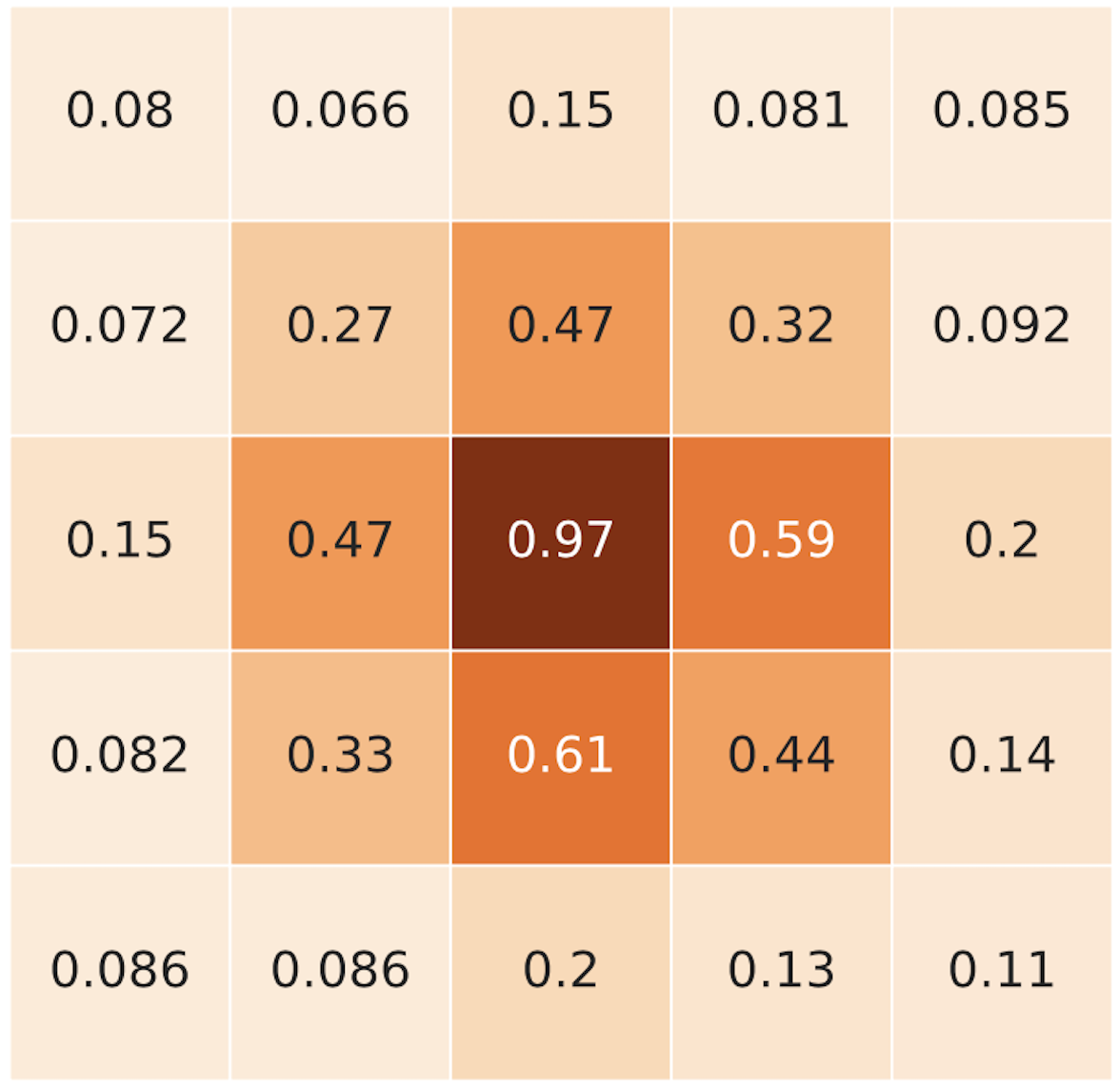}
    \caption{MNASNet}
    \label{p5-4}
  \end{subfigure}
  \caption{The average kernel magnitude matrices of MobileNetV2, MobileNetV3, EfficientNet and MNASNet trained on ImageNet.}
  \label{p5}
  \vspace{-1em} 
\end{figure}

\begin{figure}[t]
  \centering
  \includegraphics[width=0.47\textwidth]{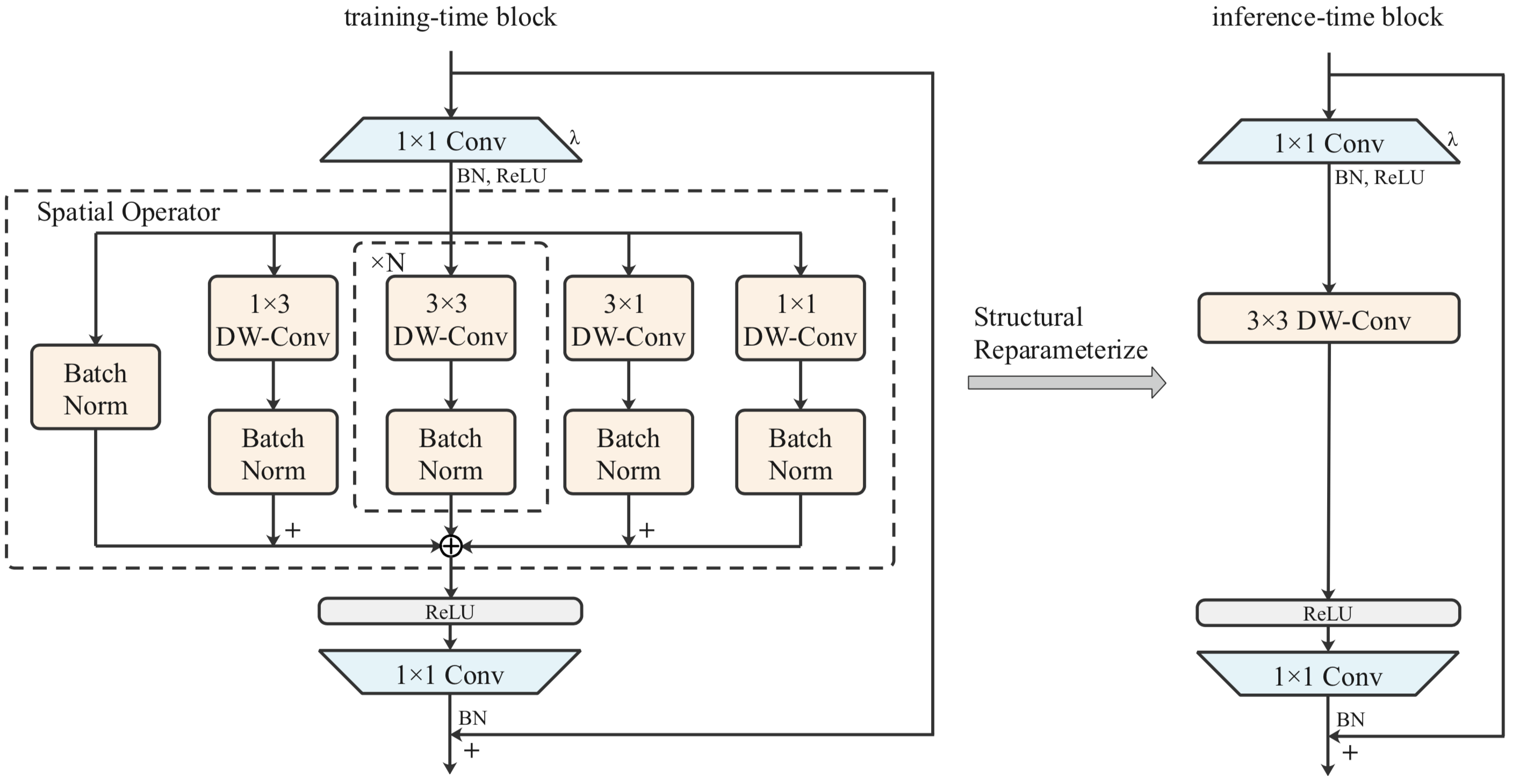}
  \caption{Meta Light Block with RepSO}
  \label{p4}
  \vspace{-1em} 
\end{figure}

\begin{figure*}[t]
  \centering
  \includegraphics[width=0.88\textwidth]{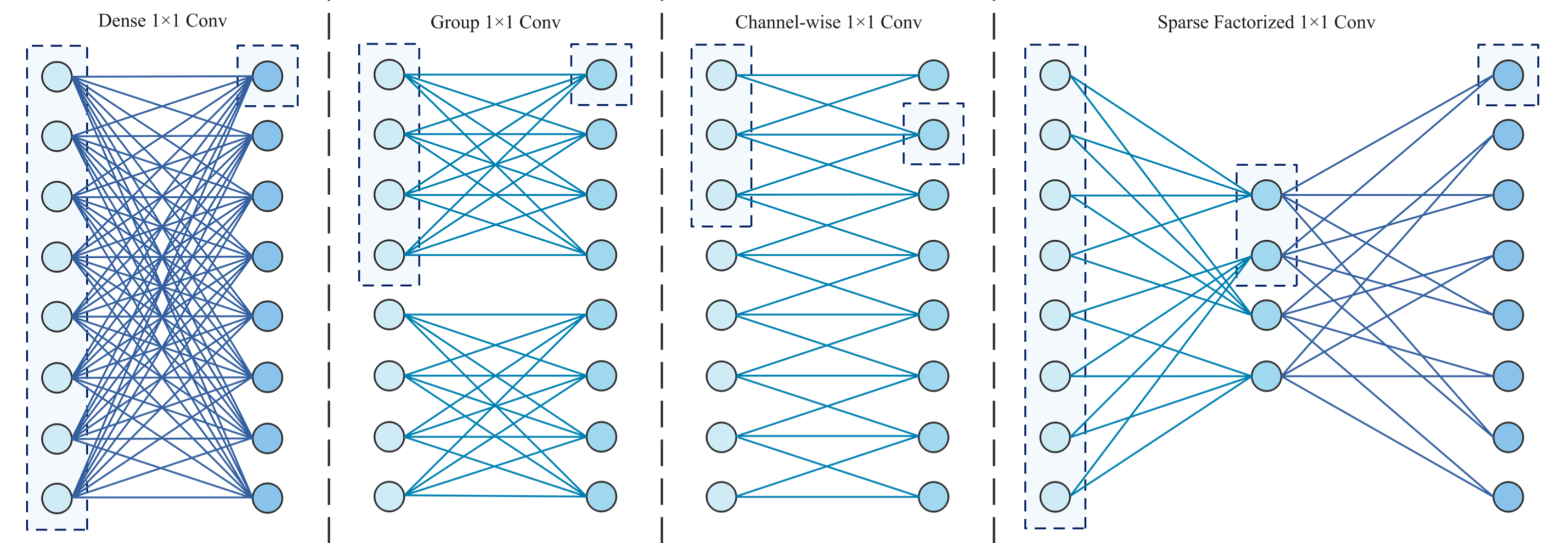}
  \caption{Connections and corresponding receptive ranges of different $1\times1$ Convs}
  \label{p7}
  \vspace{-1em} 
\end{figure*}
\subsection{Strengthen the Spatial Operator}\label{so}
Though the Meta Light Block guarantees the fundamental performance of the model with certain overall architecture, powerful spatial operator can significantly enhance the model representational capacity. 
Thus we construct multiple branches of versatile spatial operators to enrich the representation space.  
It is assumed that a position of the kernel tends to be more significant if it has a larger average kernel magnitude~\cite{ding2019acnet,ding2018auto,guo2016dynamic,han2015learning,han2015deep}. 
We first calculate and visualize the average kernel magnitude matrices of four popular light-weight CNNs as shown in Fig.~\ref{p5}. 
It is observed that the positions of the outermost circle in the $5\times5$ kernel have negligible importance compared to the central $3\times3$ positions, thus we utilize $3\times3$ convolution kernels which also have fewer parameters and Flops. To compensate for the reduced feature channels, we construct N (N=3 by default) parallel $3\times3$ DW-Conv branches.
Moreover, Fig.~\ref{p5} shows that the positions in the skeleton pattern of the $3\times3$ kernel account for much importance compared to these corner positions, thus we additionally construct horizontal $1\times3$ and vertical $3\times1$ DW-Conv branches. 
In addition, it is observed that the central position always possesses the highest importance score (almost 1), thus we construct an extra $1\times1$ DW-Conv branch to further enhance the central position.
Last but not least, as the meta block provides the fundamental capacity, we also add an identity mapping branch to the spatial operator layer.
The obtained multi-branch operator is termed as \textbf{\emph{RepSO}} (\textbf{\emph{Re}}parameterized \textbf{\emph{S}}patial \textbf{\emph{O}}perator), as sketched in Fig.~\ref{p4} left, all these seven branches are individually equipped with a batch normalization layer, then the normalized outputs of each branch are element-wise added. According to the additivity and homogeneity of convolution and the methodology of structural reparameterization~\cite{ding2019acnet}, these seven branches can be equivalently converted to a single $3\times3$ DW-Conv branch in inference as sketched in Fig.~\ref{p4} right, which produces no extra inference cost.

\subsection{Factorize the Channel Operator}
\subsubsection{Receptive Range}
For that $1\times1$ Conv accounts for large proportion of model parameters and Flops, we want to make it light-weight.
Essentially, we can change the connections between input and output neurons to change the amount of parameters. 
Hence we propose the concept of \textbf{\emph{Receptive Range}} as the guideline for establishing these connections.
The concept of receptive range is introduced specifically for channel dimension and analogous to the receptive field for spatial dimension. The value of the receptive range of a certain output neuron is the number of input neurons that it attend to directly (by a weight) or indirectly (attend to a set of hidden neurons which attend to the input neurons).
Fig.~\ref{p7} exhibits the connections and corresponding receptive ranges of different $1\times1$ Convs.
Suppose $C$ is the number of input neurons (channel numbers), as observed, each output neuron of dense $1\times1$ Conv can attend to all the input neurons directly, thus the receptive range is $C$. 
For group $1\times1$ Conv with $g$ groups, the receptive range is $\frac{C}{G}$, thus different groups can establish information connections, causing significant reduce of representation capacity.
For channel-wise $1\times1$ Conv~\cite{convolutions2021channelnets} with window size of $k$, the receptive range is $k$, thus each output neuron can only attend to small amount of input neurons, leading to insufficient channel information aggregating.
Consequently, to make the channel information aggregated sufficiently thus guarantee the model representation capacity, each output neuron should have a full receptive range, namely, each output neuron should be connected to all of the input neurons directly or indirectly.

\begin{figure*}[t]
  \centering
  \includegraphics[width=0.88\textwidth]{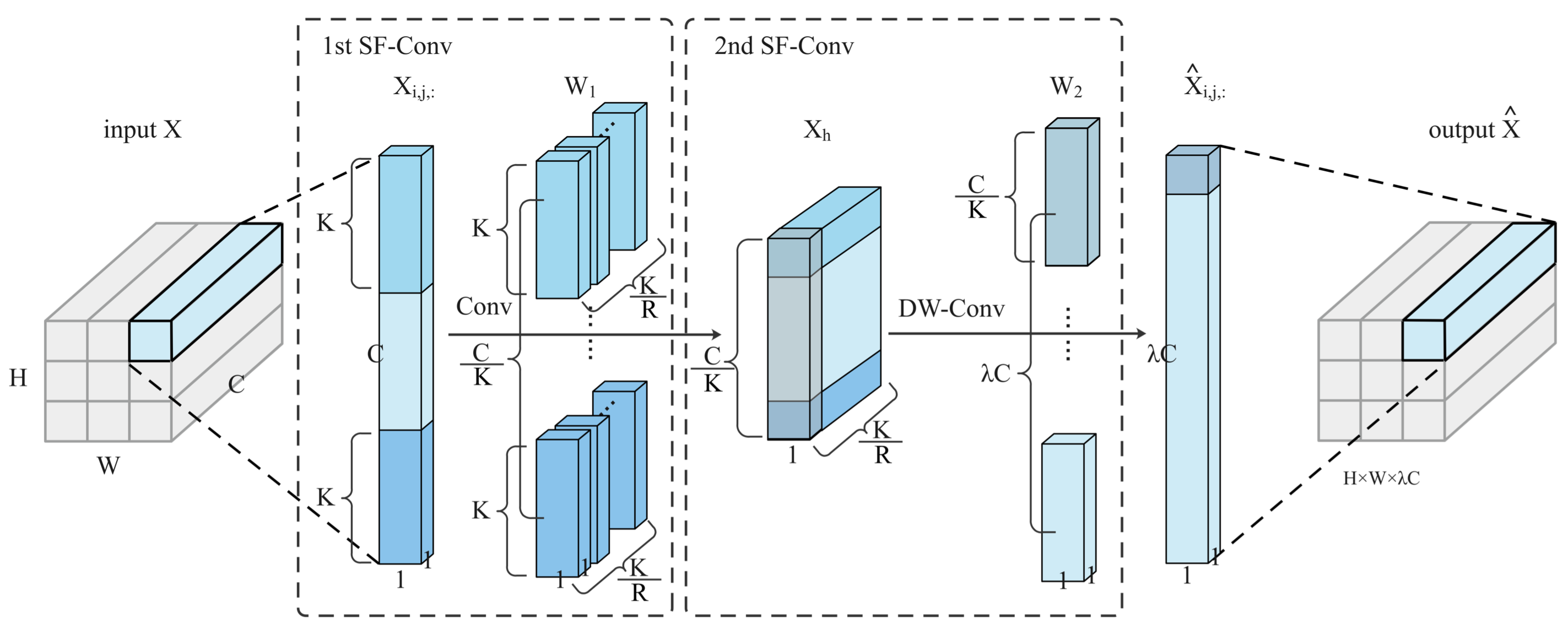}
  \caption{Sparse Factorized $1\times1$ Convolution.}
  \label{p6}
  \vspace{-1em} 
\end{figure*}

\begin{figure}[t]
  \centering
  \includegraphics[width=0.47\textwidth]{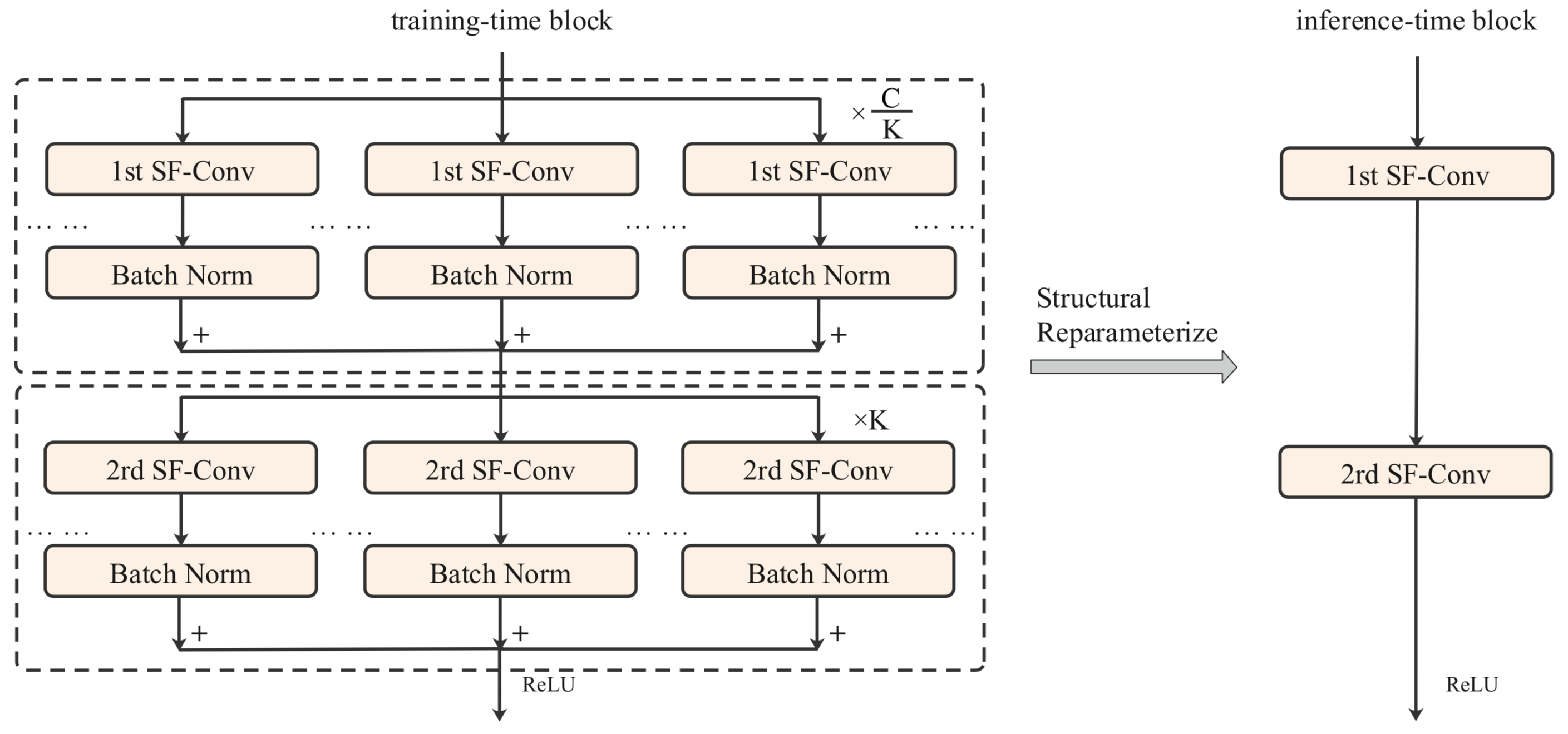}
  \caption{RefCO Channel Operator}
  \label{p8}
  \vspace{-1em} 
\end{figure}

\subsubsection{Sparsely Factorized $1\times1$ Conv}
With the guidance of full receptive range, we try to introduce sparsity to the densely connected $1\times1$ conv emulating the spatial convolution. 
We raise the \textbf{\emph{Sparsely Factorized $1\times1$ Conv}} (\textbf{\emph{SF-Conv}} for short), which is proposed to be a new paradigm of channel sparse connectivity. As Fig.~\ref{p6} exhibits, SF-Conv factorizes a densely connected $1\times1$ conv into two sparsely connected $1\times1$ convs, namely, 1st and 2nd SF-Convs, to guarantee the full receptive range. 
Given input feature map of tensor $\text{X}\in\mathbb{R}^{H\times W\times C_{in}}$, where $H, W$is the spatial size, and $C_{in}$ is the input channel numbers. The output feature map is $\hat{\text{X}}\in\mathbb{R}^{H\times W\times C_{out}}$, where $C_{out}=\lambda C_{in}=\lambda C$. 
For certain operation, we only consider the number of parameters since $Flops=H\times W\times Params$ where $H, W$ are fixed. 
For a SF-Conv , we can consider $\text{X}_{i,j,:}\in\mathbb{R}^{C\times 1\times 1}$ for each pixel $(i,j)$ in $\text{X}$ as an actual input feature map with a spatial size of ${C_{in}\times 1}$ and 1 channel, thus we can conduct standard convolution on it, which introduces sparsity connection.  
The 1st SF-Conv has a channel reduction coefficient $R$ (hyper-parameter, $R=2$ by default) to control the neuron number of the hidden feature map $\text{X}_h$ thus control the parameters. Thus there are $\frac{C}{R}$ neurons in $\text{X}_h$.
Suppose a convolution kernel $\hat{\text{W}}_1$ with spatial size of $K\times1$.
Consider the case of single channel, sliding $K\times1$ on the input $C\times1$ with stride $S$ generates $\frac{C-K+S}{S}$ output neurons. Thus to generate $\frac{C}{R}$ hidden neurons, the convolution kernel should have $\frac{C}{R}/\frac{C-K+S}{S}$ channels, that is $\text{W}_1\in\mathbb{R}^{(\frac{C}{R}/\frac{C-K+S}{S})\times 1\times K\times1}$, and the feature map of the hidden neurons is $\text{X}_h\in\mathbb{R}^{\frac{C-K+S}{S}\times1\times(\frac{C}{R}/\frac{C-K+S}{S})}$.
To achieve the minimal connections while obtain the maximum receptive range, the stride $S$ it to be $S=K$. 
Moreover, to enhance the representational capacity and increase the degree of freedom, we make the kernel weights unshared in the spatial dimension, that is, each set of input neurons is operated by a individual set of weights. 
As a consequence, the weights of 1st SF-Conv is $\text{W}_1\in\mathbb{R}^{\frac{K}{R}\times\frac{C}{K}\times K\times1}$, and the output of 1st SF-Conv is $\text{X}_h\in\mathbb{R}^{\frac{C}{K}\times1\times\frac{K}{R}}$.
Thus 1st SF-Conv has $\frac{K}{R}\times\frac{C}{K}\times K\times1=\frac{CK}{R}$ parameters.

For each output neuron of 1st SF-Conv has a receptive range of $K$, while for the set of neurons in a certain channel of $\text{X}_h$ has a total receptive range of $C$. Thus through attending to the set of neurons in a certain channel of $\text{X}_h$, the output neuron of SF-Conv can have a full receptive range with minimal number of parameters.
Thus the 2nd SF-Conv takes $\text{X}_h\in\mathbb{R}^{\frac{C}{K}\times1\times\frac{K}{R}}$ as the input and conducts a DW-Conv on it, thus the weights of 2nd SF-Conv $\text{W}_2$ has a spatial kernel size of $\frac{C}{K}\times1$. To obtain the $\lambda C$ output channels, $\text{W}_2$ should have a width multiplier of $\lambda C/\frac{K}{R}$, thus $\text{W}_2\in\mathbb{R}^{\lambda C\times1\times\frac{C}{K}\times1}$ with $\frac{\lambda C^2}{K}$ parameters.
The output feature map of the 2nd SF-Conv is $\hat{\text{X}}_{i,j,:}\in\mathbb{R}^{\lambda C\times 1\times 1}$ for the pixel $(i,j)$ in $\hat{\text{X}}$. Through operating the shared SF-Conv in all spatial positions of $\text{X}$, we can obtain the $\hat{\text{X}}$. 

SF-Conv has a total parameters of $\frac{CK}{R}+\frac{\lambda C^2}{K}$. Given a certain channel reduction ratio $R$, to achieve the least parameters, the kernel size $K$ is set as $K=\sqrt{\lambda CR}$, which is dynamically adjusted according to $C$, $\lambda$ and $R$. The total number of parameters of SF-Conv becomes $2C\sqrt{\frac{\lambda C}{R}}$.
For a dense $1\times1$ conv with weights $\text{W}_d \in\mathbb{R}^{\lambda C\times C\times1\times1}$ has $\lambda C^2$ parameters. Therefore, SF-Conv exhibits a parameter count that is only $\frac{2}{\sqrt{\lambda CR}}$ of the parameter count of PW-Conv.
In this way, SF-Conv can introduce sparsity to the channel connections, as well as maintain the full receptive range (as shown in Fig.~\ref{p7}), thus reducing the number of parameters and Flops while obtain a competitive representation capacity.

\begin{figure}[t]
  \centering
  \includegraphics[width=0.47\textwidth]{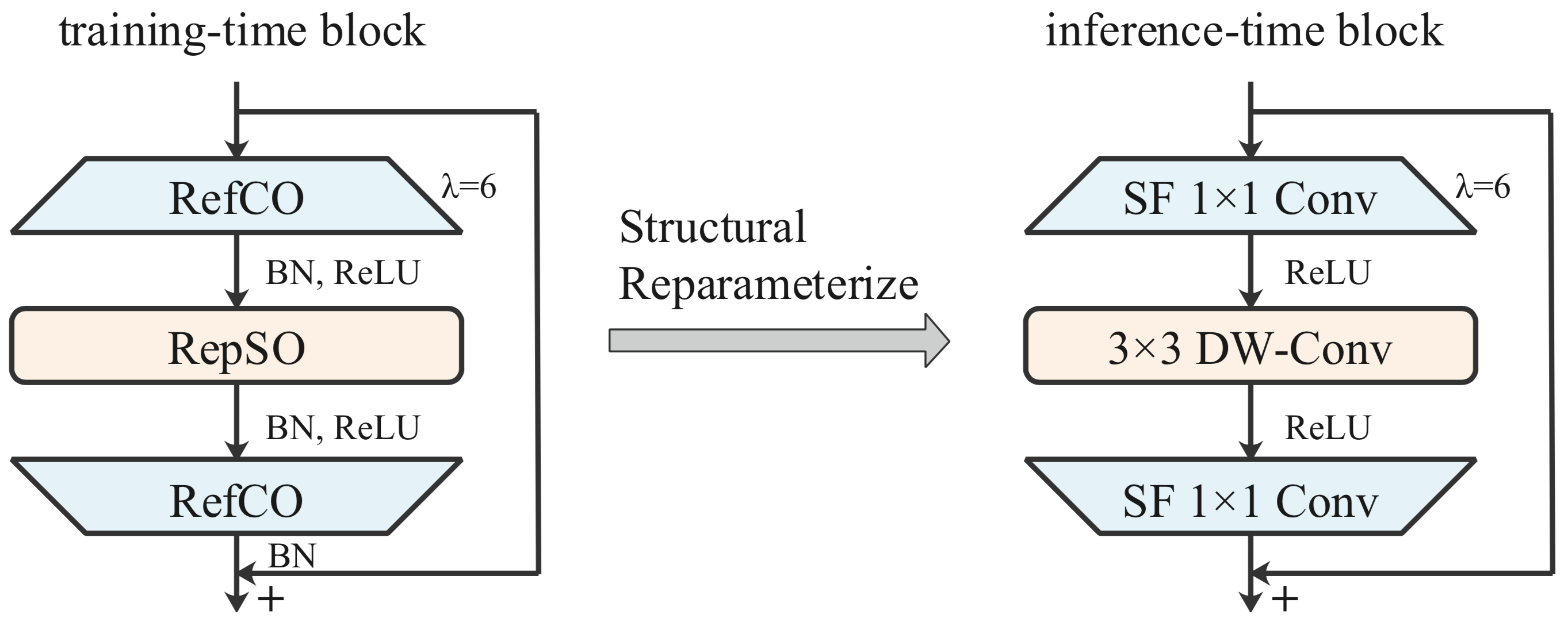}
  \caption{Meta Light Block with RepSO and RefCO for FalconNet}
  \label{p9}
  \vspace{-1em} 
\end{figure}
\subsubsection{RefCO}
We further raise \textbf{\emph{RefCO}} as the \textbf{\emph{Re}}parameterized \textbf{\emph{f}}actorized \textbf{\emph{C}}hannel \textbf{\emph{O}}perator. RefCO also employs the structural reparameterization methodology to compensate for the reduced parameter count and enhance representations. 
As Fig.~\ref{p8} exhibits, in training, RefCO firstly constructs $\frac{C}{K}$ parallel 1st SF-Conv branches and added the outputs, then constructs $K$ parallel 2nd SF-Conv branches and added the $K$ output feature maps to obtain the final output. In inference, these parallel 1st/2nd SF-Conv branches can be equivalently converted into a single SF-Conv branch.

\begin{table*}[t]
\caption{Results of various light-weight CNN models and LightNets with corresponding basic blocks on CIFAR-10, CIFAR-100 and Tiny-ImageNet-200. Width multiplier is used for some LightNets to compensate for the largely reduced channels of certain basic blocks, such as Sandglass and ShuffleNet blocks. The numbers of parameters and Flops in inference are also exhibited.}
\centering
\begin{tabular}{@{}ll|ccc|cc@{}}
\toprule
Basic Block & Model & CIFAR-10 & CIFAR-100 & Tiny-ImageNet-200 & Params & \ \ \ \ \  Flops\ \ \ \ \ \  \\
\midrule
\multirow{2}{*}{DSC Block }
&MobileNetV1&	93.18\%&	72.47\%&	64.36\%&	4.23M&	588.91M\\
\cmidrule{2-7}
&\textbf{LightNet$\times3.5$}&	93.22\%&	74.26\%&	64.38\%&	3.59M&	536.06M\\
\midrule
\multirow{3}{*}{Residual DSC Block }
&ResNet-18&	 93.89\%&	  74.52\%&	  65.02\%&	11.72M& 	1844.08M\\
&ResNet-34&	 93.95\%&	  74.82\%&	  65.21\%&	21.84M&	3698.78M\\
\cmidrule{2-7}
&\textbf{LightNet$\times3.5$}&	93.70\%&	75.01\%&	64.66\%&	3.59M&	536.06M	\\
\midrule
\multirow{2}{*}{ShuffleNetV1 Block }
&ShuffleNetV1&	 91.35\%&	  68.51\%&	  56.86\%&	1.81M&	138.75M\\
\cmidrule{2-7}
&\textbf{LightNet$\times4.0$}&	 91.61\%&	  69.42\%&	  58.94\%&	1.05M&	132.86M\\
\midrule
\multirow{2}{*}{ShuffleNetV2 Block }
&ShuffleNetV2&	 92.31\%&	  70.08\%&	  60.38\%&	2.28M&	154.87M\\
\cmidrule{2-7}
&\textbf{LightNet$\times3.5$}&	 93.50\%&	  73.56\%&	  64.10\%&	2.01M&	219.65M\\
\midrule
\multirow{6}{*}{Inverted Residual Block }
&MobileNetV2&	 93.43\%&	  74.03\%&	  66.30\%&	3.56M&	353.01M\\
&MobileNetV3S&	 91.89\%&	  70.50\%&	  60.18\%&	2.94M&	66.89M\\
&MobileNetV3L&	 94.07\%&	  73.56\%&	  65.54\%&	5.48M&	238.85M\\
&EfficientNetV1-B0&	 94.21\%&	  75.68\%&	  66.62\%&	4.98M&	404.42M\\
&EfficientNetV2-S&	 93.85\%&	  75.42\%&	  67.18\%&	21.14M&	2915.26M\\
\cmidrule{2-7}
&\textbf{LightNet}&	 94.84\%&	  76.92\%&	  68.18\%&	3.32M&	526.51M\\
\midrule
\multirow{2}{*}{Sandglass Block }
&MobileNeXt&	 93.01\%&	  67.42\%&	  58.12\%&	3.31M&	310.04M\\
\cmidrule{2-7}
&\textbf{LightNet}$\times6.0$&	 94.09\%&	  75.43\%&	  65.76\%&	3.77M&	607.85M\\
\midrule
\multirow{2}{*}{HBO Block }
&HBONet&	 92.32\%&	  72.39\%&	  64.20\%&	4.56M&	326.99M\\
\cmidrule{2-7}
&\textbf{LightNet}&	 92.44\%&	  72.85\%&	  65.92\%&	3.83M&	209.23M\\
\midrule
\multirow{2}{*}{FasterNet Block }
&FasterNet-T0&	 92.94\%&	  68.02\%&	  58.32\%&	3.64M&	310.98M\\
\cmidrule{2-7}
&\textbf{LightNet}&	 93.45\%&	  74.14\%&	  64.78\%&	3.36M&	509.10M\\
\midrule
\textbf{RepSO Block} &\textbf{LightNet}&	 94.96\%&	 78.24\%&	  69.72\%&	3.32M&	526.51M\\
\midrule
\textbf{RepSO+RefCO Block} &\textbf{FalconNet}&	 94.85\%&	 78.04\%&	  69.46\%&	2.39M&	333.14M\\

\bottomrule
\end{tabular}
\label{t1}
\end{table*}

\begin{table*}[t]
\caption{Results of different meta basic block instantiations on CIFAR-10, CIFAR-100 and Tiny-ImageNet-200.}
\centering
\begin{tabular}{@{}clllllc|cc@{}}
\toprule
Index & 1st SO & 2nd SO & 3rd SO & 1st CO & 2nd CO & Exp Ratio & CIFAR-10 & CIFAR-100\\
\midrule
A&Identity& Identity&	 Identity&	 PW-Conv &  PW-Conv &   $\lambda=6$&	 90.38\%&	67.56\%	\\
B&DW-Conv & Identity&	 Identity&	 PW-Conv &  PW-Conv &   $\lambda=4$&	 91.55\%&	-- --	\\
C&DW-Conv & Identity&	 Identity&	 PW-Conv &  PW-Conv &   $\lambda=6$&	 92.11\%&	-- --	\\
D&Identity& DW-Conv &	 Identity&	 PW-Conv &  PW-Conv &   $\lambda=4$&	 94.31\%&	-- --	\\
E&\textbf{Identity}& \textbf{DW-Conv }&	 \textbf{Identity}&   \textbf{PW-Conv }&  \textbf{PW-Conv }&   \textbf{$\lambda=6$}&	 \textbf{94.84\%}&	\textbf{76.92\%}\\
F&DW-Conv & DW-Conv &	 Identity&   PW-Conv &  PW-Conv &   $\lambda=6$&	 93.36\%&	75.10\%	\\
G&DW-Conv & DW-Conv &	 DW-Conv &   PW-Conv &  PW-Conv &   $\lambda=6$&	 91.83\%&	74.92\%	\\
H&P-Conv  & Identity&	 Identity&	 PW-Conv &  PW-Conv &   $\lambda=4$&	 92.75\%&	73.58\%	\\
I&P-Conv  & Identity&	 Identity&	 PW-Conv &  PW-Conv &   $\lambda=6$&	 93.45\%&	74.14\%	\\
J&Conv    & Identity&	 Identity&   PW-Conv &  PW-Conv &   $\lambda=4$&	 93.58\%&	73.42\%	\\
K&Conv    & Identity&	 Identity&   PW-Conv &  PW-Conv &   $\lambda=6$&	 93.67\%&	73.89\%	\\
L&Conv    & DW-Conv &	 Identity&   PW-Conv &  PW-Conv &   $\lambda=4$&	 93.71\%&	74.40\%	\\
M&Conv    & DW-Conv &	 Identity&   PW-Conv &  PW-Conv &   $\lambda=6$&	 94.10\%&	75.42\%	\\
N&DW-Conv & Identity&	 Identity&	 PW-Conv &  Identity&   $\lambda=1$&	 93.70\%&	74.26\%	\\
O&Identity& DW-Conv &	 Identity&	 PW-Conv &  PW-Conv &   $\lambda=1$&	 93.16\%&	75.01\%	\\
P&DW-Conv & DW-Conv &	 Identity&	 PW-Conv &  PW-Conv &   $\lambda=1$&	 93.29\%&	74.92\%	\\
Q&DW-Conv & Identity&	 DW-Conv &   PW-Conv &  PW-Conv &   $\lambda=1/6$&	 94.09\%&	75.43\%	\\
R&DW-Conv & DW-Conv&	 DW-Conv &   PW-Conv &  PW-Conv &   $\lambda=1/6$&	 92.11\%&	-- --	\\
S&PW-Conv & Identity&	 Identity&	 DW-Conv &  PW-Conv &   $\lambda=6$&	 92.27\%&	-- --	\\
T&PW-Conv & Identity&	 PW-Conv &	 DW-Conv &  DW-Conv &   $\lambda=6$&	 91.59\%&	-- --	\\
\bottomrule
\end{tabular}
\label{t2}
\end{table*}

\begin{table*}[t]
\caption{Results of Meta Light Block different instantiations of spatial operator on CIFAR-10, CIFAR-100 and Tiny-ImageNet-200. Each row exhibits the branch numbers of DW-Conv with certain kernel size that consist the corresponding spatial operator.}
\centering
\begin{tabular}{@{}cccccccccccc|cc@{}}
\toprule
Index&Identity & $1\times1$  & $1\times3$  & $3\times1$  & $3\times3$   & $1\times5$  & $5\times1$ & $3\times5$  & $5\times3$ & $5\times5$ & $7\times7$& CIFAR-10 & CIFAR-100 \\
\midrule
A&0& 0&  0&	 0&  0&  0&  0&  0&	 0&  0&	 0&    90.38\%&	 68.56\%\\
B&0& 0&  1&	 1&  0&  0&  0&  0&	 0&  0&	 0&    94.86\%&	 76.85\%\\
C&0& 0&  0&	 0&  1&  0&  0&  0&	 0&  0&	 0&    94.84\%&	 76.92\%\\
D&0& 0&  0&	 0&  0&  1&	 1&  0&  0&  0&	 0&    94.16\%&	 76.20\%\\
E&0& 0&  0&	 0&  0&  0&	 0&  0&  0&  1&	 0&    94.31\%&	 76.27\%\\
F&0& 0&  0&	 0&  0&  0&	 0&  0&  0&  0&	 1&    93.94\%&	 75.21\%\\
G&1& 0&  0&	 0&  1&  0&	 0&  0&  0&  0&	 0&    94.64\%&	 76.10\%\\
H&0& 1&  0&	 0&  1&  0&	 0&  0&  0&  0&	 0&    94.61\%&	 76.88\%\\
I&0& 0&  1&	 1&  1&  0&	 0&  0&  0&  0&	 0&    94.67\%&	 77.24\%\\
J&0& 0&  0&	 0&  1&  1&	 1&  0&  0&  0&	 0&    94.59\%&	 76.54\%\\
K&0& 0&  0&	 0&  1&  0&	 0&  1&  1&  0&	 0&    94.63\%&	 76.42\%\\
L&0& 0&  0&	 0&  1&  0&	 0&  0&  0&  1&	 0&    94.75\%&	 76.27\%\\
M&0& 1&  1&	 1&  1&  0&	 0&  0&  0&  0&	 0&     -- -- &	 77.58\%\\
N&0& 0&  1&	 1&  1&  1&	 1&  0&  0&  1&	 0&     -- -- &	 76.98\%\\
O&0& 1&  1&	 1&  1&  1&	 1&  1&  1&  1&	 0&     -- -- &	 76.92\%\\
P&0& 0&  3&	 3&  0&  0&	 0&  0&  0&  0&	 0&     -- -- &	 76.01\%\\
Q&0& 0&  0&	 0&  3&  0&	 0&  0&  0&  0&	 0&     -- -- &	 77.69\%\\
R&0& 0&  1&	 1&  3&  0&	 0&  0&  0&  0&	 0&     -- -- &	 77.92\%\\
S&0& 1&  1&	 1&  3&  0&	 0&  0&  0&  0&	 0&    94.92\%&	 78.23\%\\
T&1& 1&  1&	 1&  3&  0&	 0&  0&  0&  0&	 0&    94.96\%&	 78.24\%\\
\bottomrule
\end{tabular}
\label{t3}
\end{table*}

\subsection{FalconNet}
Based on above four vital components, namely, LightNet overall architecture, Meta Light Block, RepSO spatial operator and RefCO channel operator, we obtain a novel light-weight CNN model termed as \textbf{\emph{FalconNet}} (\textbf{\emph{Fa}}ctorization for the \textbf{\emph{l}}ight-weight \textbf{\emph{con}}v\textbf{\emph{Net}}). 
Fig.~\ref{p8} exhibits the Meta Light Block with RepSO and RefCO, which is utilized as the basic block for FalconNet. 
Later experimental results validate that FalconNet can achieve higher accuracy with lower number of parameters and Flops compared to existing light-weight CNNs. 

\section{Experiments}
\subsection{Configurations}
We conduct abundant experiments on three challenging benchmark datasets, CIFAR-10, CIFAR-100 and Tiny-ImageNet-200, to validate the effectiveness and superiority of the four vital components illustrated above.
CIFAR-10/100 consists of 50K training images and 10K testing images, Tiny-ImageNet-200 contains 100K training images and 10K testing images.
For the training configuration, we use the cross entropy loss function and adopt an SGD optimizer with momentum of 0.9, batch size of 256, and weight decay of $4\times10^{-5}$, as the common practice~\cite{ding2022scaling,sandler2018mobilenetv2}. We use a learning rate schedule with a 5-epoch warmup, initial value of 0.1, and cosine annealing for 300 epochs to guarantee the compete convergence. The data augmentation uses random cropping and horizontal flipping. The input resolution is uniformly resized to $224\times224$.
All the models are random initialized with Xavier initialization and trained with the same training configuration from scratch.

\subsection{Performance Evaluation}\label{exp-lightnet}
We evaluate the performance of various existing light-weight CNN models, including MobileNetV1/V2/V3~\cite{howard2017mobilenets,sandler2018mobilenetv2,howard2019searching}, MobileNeXt~\cite{zhou2020rethinking}, EfficientNetV1/V2~\cite{tan2019efficientnet,tan2021efficientnetv2}, ShuffleNetV1/V2~\cite{zhang2018shufflenet,ma2018shufflenet}, HBONet~\cite{li2019hbonet}, and FasterNet~\cite{chen2023run}, as well as two heavy-weight CNNs, i.e., ResNet-18 and ResNet-34~\cite{he2016deep}.
Then we implement the basic blocks of these light-weight CNN models to our LightNet overall architecture and compare the performance with existing light-weight CNNs. Table~\ref{t1} shows the experimental results.
As can be observed, LightNet can achieve better performance by simply implementing the basic blocks.
For example, LightNet with Inverted Residual Block surpasses EfficientNetV1 by 0.63\%, 1.24\% and 1.56\% on CIFAR-10/100 and Tiny-ImageNet-200 respectively with only 66\% of the parameters (there exists a trade-off that the Flops is enhanced by 30\%). Moreover, LightNet with ShuffleNetV2 Block significantly surpasses ShuffleNetV2 by 3.48\% and 3.72\% on CIFAR-100 and Tiny-ImageNet-200 with 88\% parameters (trade-off of 40\% more Flops). Besides, LightNet with Sandglass Block significantly surpasses MobileNeXt by 8.01\% and 9.94\% on CIFAR-100 and Tiny-ImageNet-200 with 14\% more parameters, and LightNet with FasterNet Block surpasses FasterNet by 6.08\% and 6.46\% on CIFAR-100 and Tiny-ImageNet-200 with 8\% less parameters. 
In addition, when compared to the heavy-weight CNN models ResNet-18/34, LightNet with Residual DSC Block (obtained by replacing the standard convolution in the bottleneck block of ResNet with DS-Conv) still achieves competitive performance, while the numbers of parameters are reduced by 70\% and 84\% respectively.
The experimental results validate the effectiveness and efficiency of LightNet overall architecture compared to the overall architectures of existing light-weight CNN models.

Then we evaluate the performance of LightNet with RepSO Block (Meta Light Block with RepSO), it is observed in Table~\ref{t1} that RepSO can enhance the performance of LightNet significantly and achieves the highest accuracy on all of the three datasets, for example, it surpasses LightNet with Inverted Residual Block by 1.32\% and 1.54\% on CIFAR-100 and Tiny-ImageNet-200 respectively, while maintaining the inference parameters and Flops unchanged. Thus validates the effectiveness of RepSO which boosts the morel performance significantly without incurring any extra inference costs. 

We then evaluate the performance of FalconNet, which equips LightNet with Meta Light Block of RepSO and RefCO.
As Table~\ref{t1} exhibits, FalconNet can achieve higher accuracy than other existing light-weight CNNs while possessing less parameters and Flops. Moreover, compared to FalconNet without RefCO (namely, LightNet with RepSO Block), FalconNet significantly reduces the number of parameters and Flops by 28\% and 37\% while still achieving competitive accuracy with negligible decline of 0.11\%, 0.20\% and 0.26\% on CIFAR-10, CIFAR-100 and Tiny-ImageNet-200 respectively.
This validates the effectiveness of RefCO which significantly reduce the number of parameters and Flops while maintaining good representation capacity and competitive performance.

\subsection{Ablation Study}
\subsubsection{Meta Basic Block}\label{exp-metablock}
We first conduct various ablation study to find some principles in instantiating the Meta Basic Blocks, Table~\ref{t2} exhibits the experimental results.
We test different combinations of the 1st/2nd/3rd spatial operators and the 1st/2nd channel operators (with the expansion ratio $\lambda$).
The \emph{A} experiment validates that Meta Basic Block provides the basic ability for the model since the model when having no learnable parameters for spatial operators (i.e., identity) still achieve satisfying performance.
Compare the experiments of \emph{E,F,G,M}, it can be concluded that instantiating the 1st/3rd spatial operators with DW-Conv will undermine the final performance. 
Moreover, compare the experiments of \emph{C,E,I,K,S,T}, it can be concluded that instantiating the 1st/3rd spatial operators learnable while instantiating the wnd spatial operator identity will undermine the final performance.
Besides, experiments \emph{N,S,T} illustrate that these two channel operators of PW-Convs are significant.
Thus we further simplify the Meta Basic Block into Meta Light Block as illustrated in Sec.~\ref{mb} and Fig.~\ref{p3-2-1}, which only maintain the 2nd spatial operator and 1st/2nd channel operator while leaving the other two spatial operators.

\subsubsection{Spatial Operator}\label{exp-sp}
Here we conduct ablation studies to explore the form of the spatial operator in the Meta Light Block, Table~\ref{t3} exhibits the experimental results. Each row exhibits the branch numbers of DW-Conv with certain kernel size that consist the corresponding spatial operator.
From experiments \emph{C,E,F}, it can be observed that employing large convolutional kernels does not contribute to the improvement of model representational capacity. Instead, it adversely impacts the model's performance. This conclusion is consistence with the phenomenon stated in Sec.~\ref{so} and Fig.~\ref{p5}.
In addition, experiments \emph{B,C,I,S,T} validate that adding horizontal and vertical branches can enhance the representational capacity as stated in Sec.~\ref{so} and Fig.~\ref{p5}.

\section{Conclusion}
This paper factorizes the structure of light-weight from coarse to fine and obtain four vital components, namely, overall structure, basic block, spatial operator and channel operator. In light of the existing issues of these components, this paper respectively raise LightNet overall architecture, Meta Light Block, RepSO spatial operator, concept of receptive range and RefCO channel operator. Based on these four components, this paper raises FalconNet, which achieves higher accuracy with lower number of parameters and Flops compared to existing light-weight CNNs.


\end{document}